\documentclass[conference]{IEEEtran}
\usepackage{amsmath,amssymb,amsthm,epsfig,dsfont,color,subfigure,empheq,graphicx}
\usepackage{enumerate,stfloats,url,algorithm,algorithmic}
\DeclareMathOperator{\diag}{diag}

\newtheorem{lemma}{Lemma}

\theoremstyle{remark}

\begin{document}
\title{Grid Topology Identification using Electricity Prices}


\author{
\authorblockN{Vassilis Kekatos and Georgios B. Giannakis}
\authorblockA{Digital Tech. Center and Dept. of ECE, Univ. of Minnesota\\
Minneapolis, MN 55455\\
Emails: \{kekatos,georgios\}@umn.edu}
\and
\authorblockN{Ross Baldick}
\authorblockA{Dept. of ECE, Univ. of Texas at Austin\\
Austin, TX 78712\\
Email: baldick@ece.utexas.edu}
}

\maketitle

\begin{abstract}
The potential of recovering the topology of a grid using solely publicly available market data is explored here. In contemporary whole-sale electricity markets, real-time prices are typically determined by solving the network-constrained economic dispatch problem. Under a linear DC model, locational marginal prices (LMPs) correspond to the Lagrange multipliers of the linear program involved. The interesting observation here is that the matrix of spatiotemporally varying LMPs exhibits the following property: Once premultiplied by the weighted grid Laplacian, it yields a low-rank and sparse matrix. Leveraging this rich structure, a regularized maximum likelihood estimator (MLE) is developed to recover the grid Laplacian from the LMPs. The convex optimization problem formulated includes low rank- and sparsity-promoting regularizers, and it is solved using a scalable algorithm. Numerical tests on prices generated for the IEEE 14-bus benchmark provide encouraging topology recovery results.
\end{abstract}


\section{Introduction}\label{sec:intro}
\renewcommand{\thefootnote}{\fnsymbol{footnote}}
\footnotetext{Work in this paper was supported by the Inst. of Renewable Energy and the Environment (IREE) under grant no. RL-0010-13, Univ. of Minnesota, and NSF Grant ECCS-1202135.}

Data analytics is a major driver towards the smart grid transformation: from load and wind forecasts, to consumer preference learning, and cyber-physical attack detection. Among grid data mining tasks, topology inference is one of vital importance. Currently, system operators and utilities maintain a detailed physical model for the underlying transmission and distribution grids. The model is updated regularly by the network topology processor, and constitutes the foundation for several monitoring and market-related tasks; see e.g., \cite{AburExpositoBook}.

On top of these conventional cases, identifying the grid topology can be useful for various purposes. A well-designed cyber attack on the state estimator requires grid topology information. Knowing the power network structure could assist in informed bidding strategies, or even lead to market manipulation; see e.g., \cite{LeHuBaPi11}. In addition, the Laplacian matrix of the graph corresponding to the grid provides a meaningful notion of inter-bus electrical distances. Such distances could surrogate the spatial correlation among pricing nodes \cite{KZG14}; or adopted by clustering techniques to reveal influential nodes.

Albeit the extensive research on data attacks, grid topology recovery using readily accessible data has been overlooked. The possibility of arbitrarily perturbing power system state estimates by counterfeiting a few measurements is recognized in \cite{LiNi11}. The strength of such data attacks and their impact on market outcomes is characterized in \cite{KoJi11}, \cite{LangTong14}; yet designing stealth attacks oftentimes assumes that the attacker knows the topology of the grid \cite{XiMo11}. From the system operator's point of view, detecting topological changes in the form of transmission line outages has been studied in \cite{abur10naps}, \cite{tate09pesgm}, \cite{ZhGi12}. Given hourly topology updates and nodal voltage phases, the sparse overcomplete representation employed in \cite{ZhGi12} can reveal outages of even geographically distant lines at affordable complexity. In a microgrid scenario, the timing difference of power line communication signals to travel between electrical nodes is utilized to recover the grid structure in \cite{ErTpVi13}. Grid topology recovery is cast as a blind identification problem in \cite{LiPoSc13}. Although \cite{LiPoSc13} exploits the positive semidefiniteness and the sparsity of the associated weighted grid Laplacian, it presumes that bus voltage phases are linearly independent and nodal injections are available across the \emph{entire} interconnection. Under a Markov random field (MRF) assumption, transmission network faults are localized using the sample covariance matrix of nodal voltage phases in \cite{HeZhang2011}. Likewise, if nodal voltage magnitudes constitute an MRF, the topology of a distribution grid is pursued by means of the sample covariance matrix of voltage magnitudes \cite{BoSch13}.

Distinct from the previous setups, this work considers grid topology recovery using publicly available market data. Real-time locational marginal prices (LMPs) are calculated based on the Lagrange multipliers of the network-constrained economic dispatch (Section~\ref{sec:problem}). The fresh idea here is that LMPs could unveil grid topology information. Specifically, let $\mathbf{L}$ be the $N\times T$ matrix collecting the real-time LMPs announced over a network of $N$ nodes over $T$ successive market clearing intervals. Interestingly, $\mathbf{L}$ satisfies the model $\mathbf{B}\mathbf{L}=\mathbf{S}$, where $\mathbf{B}$ is a positive definite and sparse matrix having non-positive off-diagonal entries, and $\mathbf{S}$ is a sparse and low-rank matrix (Section~\ref{sec:method}). Recognizing such a rich structure is the first contribution of this paper. Inspired by recent advances in learning using sparse and low-rank models, a novel recovery scheme is devised, and constitutes the second contribution. As a third contribution, an efficient algorithm based on the alternating direction method of multipliers (ADMM) is developed (Section~\ref{sec:algorithms}). Numerical tests on market data generated on the IEEE 14-bus benchmark corroborate the validity of our findings (Section~\ref{sec:simulations}).


\emph{Notation.} Lower- (upper-) case boldface letters denote column vectors (matrices); while $\mathbf{1}$ $(\mathbf{0})$ denotes the all-ones (all-zeros) vector, and $\mathbf{I}$ the identity matrix. Prime stands for matrix transposition; $\mathbf{A}\bullet\mathbf{B}$ is the matrix inner product; $\mathbf{A}\odot\mathbf{B}$ is the entrywise (Hadamard) matrix product; $\diag(\{a_i\})$ is the diagonal matrix having scalars $\{a_i\}$ on its diagonal; and $|\mathbf{A}|$ is the matrix determinant. Symbol $\mathbb{S}^N$ ($\mathbb{S}_{+}^N$) is reserved for the set of real $N\times N$ symmetric (positive semidefinite) matrices. Regarding matrix norms, $\|\mathbf{A}\|_*$ is the nuclear norm (sum of matrix singular values); $\|\mathbf{A}\|_F$
is the Frobenius norm; and $\|\mathbf{A}\|_1:=\sum_{m,n} |[\mathbf{A}]_{m,n}|$ is the $\ell_1$-norm. 

\section{Data Modeling and Problem Statement}\label{sec:problem}
Consider a power grid represented by the graph $\mathcal{G}=(\mathcal{V},\mathcal{E})$, where the set of nodes $\mathcal{V}$ corresponds to $N+1$ buses, and the edges in $\mathcal{E}$ to $L$ transmission lines. Grid connectivity is captured via the branch-bus incidence matrix $\tilde{\mathbf{A}}\in\mathbb{R}^{L\times (N+1)}$: if its $l-$th row $\tilde{\mathbf{a}}_l'$ corresponds to edge $(m,n)$, then $[\tilde{\mathbf{a}}_l]_m=+1$, $[\tilde{\mathbf{a}}_l]_n=-1$, and zero otherwise \cite{SPM2013}. For a connected grid, the nullity of $\tilde{\mathbf{A}}$ is one; and by construction $\tilde{\mathbf{A}}\mathbf{1} = \mathbf{0}$. 

To describe real-time market clearing, the linear DC power flow model is summarized next. Let $\theta_n$ be the voltage phase and $p_n$ the active power injected at bus $n$. The active power flowing from bus $m$ to bus $n$ over line $l$ is approximated as $f_l=(\theta_m-\theta_n)/x_l$, where $x_l>0$ is the line reactance. Collecting $\{\theta_n,p_n\}_{n=1}^N$ and $\{f_l\}_{l=1}^L$ in $\tilde{\boldsymbol{\theta}},~\tilde{\mathbf{p}}\in\mathbb{R}^{N+1}$ and $\mathbf{f}\in\mathbb{R}^L$, respectively, it holds that $\mathbf{f}=\mathbf{D}\tilde{\mathbf{A}}\tilde{\boldsymbol{\theta}}$, where $\mathbf{D}:=\diag\left(\{x_l^{-1}\}_{l=1}^L\right)$. Power conservation dictates that $\tilde{\mathbf{p}} = \tilde{\mathbf{A}}'\mathbf{f} = \tilde{\mathbf{B}}\boldsymbol{\tilde{\theta}}$ with $\mathbf{\tilde{B}}:=\mathbf{\tilde{A}}'\mathbf{D}\mathbf{\tilde{A}}$ defining the weighted Laplacian of $\mathcal{G}$. As such, $\mathbf{\tilde{B}}$ is positive semidefinite, and for a connected grid, its zero eigenvalue has multiplicity one with $\mathbf{1}$ being the corresponding eigenvector, i.e., $\mathbf{\tilde{B}}\mathbf{1}=\mathbf{0}$; see e.g.,~\cite{SPM2013}.

Real-time electricity prices are determined by solving the network-constrained economic dispatch problem~\cite{KirschenStrbac}. In a simple, yet sufficiently representative form, the latter is typically formulated as the linear program
\begin{subequations}\label{eq:dcopf}
\begin{align}
(\mathbf{\tilde{p}}^*,\boldsymbol{\tilde{\theta}}^*)\in\arg\min_{\tilde{\mathbf{p}},\boldsymbol{\tilde{\theta}}} ~&~\mathbf{c}'\mathbf{\tilde{p}}\label{eq:dcopf0}\\ 
\textrm{s.to}~&~\underline{\mathbf{p}}\leq \mathbf{\tilde{p}}\leq \overline{\mathbf{p}}\label{eq:dcopf1}\\
~&~ \mathbf{\tilde{p}}=\mathbf{\tilde{B}}\boldsymbol{\tilde{\theta}}\label{eq:dcopf2}\\
~&-\overline{\mathbf{f}}\leq \mathbf{D}\mathbf{\tilde{A}}\boldsymbol{\tilde{\theta}} \leq \overline{\mathbf{f}}\label{eq:dcopf3}
\end{align}
\end{subequations}
which is solved regularly to determine the (incremental) power schedules $\mathbf{\tilde{p}}^*$ for the five-minute interval ahead. 

If the $n$-th bus bidder is a generator, $\tilde{p}_n$ is non-negative and bounded by \eqref{eq:dcopf1} in accordance to generation limits and selling offers. Moreover, $[\mathbf{c}]_n$ in \eqref{eq:dcopf0} is the lowest price in dollars per MWh the generator is willing to be paid to inject power in bus $n$. If the $n$-th bus bidder is a consumer, $\tilde{p}_n$ is non-positive and it is either fixed to the value $[\underline{\mathbf{p}}]_n=[\overline{\mathbf{p}}]_n$ predicted for a fixed load, or, it is bounded by the buying offer for elastic loads. When the load is elastic, the corresponding $[\mathbf{c}]_n$ is the highest price the consumer is willing to pay for withdrawing electricity at bus $n$; or zero for fixed loads. Zero-injection buses are modeled by simply setting $\tilde{p}_n=0$ in \eqref{eq:dcopf1}. Having multiple bidders and/or multi-block bids per bus does not harm the generality of the ensuing results.

The outcome of \eqref{eq:dcopf} depends not only on participant bids $(\mathbf{c},\underline{\mathbf{p}},\overline{\mathbf{p}})$, but also on the underlying physical system limitations. Indeed, bus injections are related to nodal voltage phases via \eqref{eq:dcopf2}, while power flows $\mathbf{f}$ cannot exceed the transmis\-sion capacity limits $\overline{\mathbf{f}}$ imposed on both flow directions via \eqref{eq:dcopf3}.

The optimization in \eqref{eq:dcopf} can be simplified after recognizing that due to the nullspace of $\mathbf{\tilde{A}}$, the vector pairs $\{(\mathbf{\tilde{p}}^*,\boldsymbol{\tilde{\theta}}^* + \beta\mathbf{1}):\beta\in\mathbb{R}\}$ are minimizers of \eqref{eq:dcopf} too. This phase shift ambiguity is resolved by fixing $\theta_1=0$, in which case bus 1 serves the role of a reference bus. Consider the partition $\mathbf{\tilde{p}}:=[p_1~\mathbf{p}']'$ and $\boldsymbol{\tilde{\theta}}=[\theta_1~\boldsymbol{\theta}']'$; and let $\mathbf{A}$ be the $L\times N$ full-column rank matrix obtained after maintaining all but the first column of $\mathbf{\tilde{A}}$. Define also the \emph{reduced grid Laplacian} $\mathbf{B}:=\mathbf{A}'\mathbf{D}\mathbf{A}$, which is strictly positive definite.

The set described by \eqref{eq:dcopf2}, \eqref{eq:dcopf3}, and $\theta_1=0$, can be represented by $\mathbf{\tilde{p}}'\mathbf{1}=0$, $\mathbf{p}=\mathbf{B}\boldsymbol{\theta}$, and $\theta_1=0$. Upon eliminating $\boldsymbol{\theta}$, the problem in \eqref{eq:dcopf} can be rewritten as
\begin{subequations}\label{eq:Bdcopf}
\begin{align}
\mathbf{\tilde{p}}^*\in\arg\min_{\mathbf{\tilde{p}}} ~&~\mathbf{c}'\mathbf{\tilde{p}}\label{eq:Bdcopf0}\\ 
\textrm{s.to}~&~\underline{\mathbf{p}}\leq \mathbf{\tilde{p}}\leq \overline{\mathbf{p}}\label{eq:Bdcopf1}\\
~&~ \mathbf{\tilde{p}}'\mathbf{1}=0\label{eq:Bdcopf2}\\
~&-\overline{\mathbf{f}}\leq \mathbf{D}\mathbf{A}\mathbf{B}^{-1}\mathbf{p} \leq \overline{\mathbf{f}}.\label{eq:Bdcopf3}
\end{align}
\end{subequations}

To describe the pricing mechanism, let $\lambda_0$ denote the optimal Lagrange multiplier associated with constraint \eqref{eq:Bdcopf2}; and $\underline{\boldsymbol{\mu}}\in\mathbb{R}^{L}_{+}$ and $\overline{\boldsymbol{\mu}}\in\mathbb{R}^{L}_{+}$ be the optimal Lagrange multipliers corresponding to the lower and upper limits of \eqref{eq:Bdcopf3}, respectively. The vector of ex-ante real-time LMPs is
\begin{equation}\label{eq:lmps}
\left[\begin{array}{c}
\lambda_1\\
\boldsymbol{\lambda}
\end{array} \right]
:= \lambda_0\mathbf{1} +
\left[\begin{array}{c}
0\\
\mathbf{B}^{-1}\mathbf{A}'\mathbf{D}\boldsymbol{\mu}
\end{array} \right]+\mathbf{w}
\end{equation} 
where $\boldsymbol{\mu}:=\underline{\boldsymbol{\mu}}-\overline{\boldsymbol{\mu}}$, and $\mathbf{w}$ is a relatively small term accounting for the heat losses on transmission lines. Oftentimes, $\mathbf{w}$ is approximated as the product of $\lambda_0$ times the gradient of the system-wide loss evaluated at $\mathbf{\tilde{p}}^*$.

Evidenced by \eqref{eq:lmps}, LMPs consist of three components:\\
\hspace*{1.5em}\textbf{(C1)} the marginal energy component $\lambda_0$;\\
\hspace*{1.5em}\textbf{(C2)} the marginal congestion component; and,\\
\hspace*{1.5em}\textbf{(C3)} the marginal loss component $\mathbf{w}$.\\
Given that price components are usually announced separately, our focus will be on the topology-related information that can be extracted from (C2) and (C3). Even if LMPs are announced as a sum, (C1) can be easily subtracted from $\boldsymbol{\lambda}$ since $\lambda_1 \approx \lambda_0$, and (C3) can be modeled as low-variance noise.

As mentioned earlier, the problem in \eqref{eq:dcopf} or \eqref{eq:Bdcopf} is solved every five minutes. The grid topology captured by $\mathbf{A}$, $\mathbf{D}$, and $\mathbf{B}$, will be assumed approximately constant over a period of $T$ such intervals. However, the triplets $(\mathbf{c},\underline{\mathbf{p}},\overline{\mathbf{p}})$ are time-varying: generator offers can be altered hourly and load demands fluctuate on a 5-min basis. Let vector $\boldsymbol{\lambda}_t\in\mathbb{R}^{N}$ denote the LMPs $\boldsymbol{\lambda}$ derived for a specific 5-min interval indexed by $t$. With a slight abuse of notation, $\boldsymbol{\lambda}_t$ comprises only the (C2) plus (C3) LMP components assuming that (C1) has been removed. Due to \eqref{eq:lmps} and for $t=1,\ldots, T$, it holds that
\begin{equation}\label{eq:lmps2}
\boldsymbol{\lambda}_t =\mathbf{B}^{-1}\mathbf{A}'\mathbf{D}\boldsymbol{\mu}_t + \mathbf{n}_t=\mathbf{B}^{-1}\mathbf{s}_t+ \mathbf{n}_t
\end{equation} 
where $\mathbf{s}_t:=\mathbf{A}'\mathbf{D}\boldsymbol{\mu}_t$ and $\mathbf{n}_t$ captures unmodeled physical system variations and the loss component. Upon collecting prices and Lagrange multiplier differences over $T$ time periods in $\mathbf{L}:= [\boldsymbol{\lambda}_1~\cdots ~\boldsymbol{\lambda}_T]$ and  $\mathbf{M}:=[\boldsymbol{\mu}_1~\cdots~\boldsymbol{\mu}_T]$, model \eqref{eq:lmps2} reads
\begin{equation}\label{eq:model}
\mathbf{L} =\mathbf{B}^{-1}\mathbf{S} + \mathbf{N}
\end{equation} 
where $\mathbf{S}:=\mathbf{A}'\mathbf{D}\mathbf{M}$ and $\mathbf{N}:=[\mathbf{n}_1~\cdots~\mathbf{n}_T]$. Given $\mathbf{L}$ in \eqref{eq:model}, our goal is to identify the topology (Laplacian) matrix $\mathbf{B}$.

\section{Market Data Factorization}\label{sec:method}
Finding $\mathbf{B}$ from \eqref{eq:model} given only the observed prices in $\mathbf{L}$ constitutes a blind recovery problem, yet with a very rich structure as recognized next. According to \eqref{eq:lmps}, the $t$-th column of $\mathbf{M}$ is $\boldsymbol{\mu}_t = \underline{\boldsymbol{\mu}}_t - \overline{\boldsymbol{\mu}}_t$. Due to complementary slackness in \eqref{eq:Bdcopf}, the $l$-th entry of $\underline{\boldsymbol{\mu}}_t$ ($\overline{\boldsymbol{\mu}}_t$) is strictly positive only when line $l$ is congested, i.e., it has reached its lower (upper) capacity limit. Since typically only a few transmission lines are congested over a period of $T$ intervals, $\boldsymbol{\mu}_t$ is expected to be sparse.

The sparsity of $\mathbf{M}$ endows $\mathbf{S}$ with two important properties. Note that the $t$-th column of $\mathbf{S}$ can be expressed as $\mathbf{s}_t=\sum_{l=1}^L x_l^{-1}\mathbf{a}_l [\boldsymbol{\mu}_t]_l$. Hence, the $n$-th entry of $\mathbf{s}_t$ is non-zero only if at least one of the lines adjacent to node $n$ is congested at time $t$. Assuming that the entry $[\boldsymbol{\mu}_t]_l$ is nonzero only for a few $l\in\mathcal{E}$ across all $t$, vectors $\{\mathbf{s}_t\}_{t=1}^T$ are expected to be sparse too. Moreover, the columns of $\mathbf{S}$ are expressed by linearly combining only a few of $\{x_l^{-1}\mathbf{a}_l\}_{l\in\mathcal{E}}$. Thus, $\mathbf{S}$ is not only sparse, but low-rank as well. 

By definition of the original graph Laplacian $\mathbf{\tilde{B}}$, its $(m,n)$-th off-diagonal entry is $-x_l^{-1}$ if $m\neq n$, and 0 otherwise. Granted that power grids are sparingly connected, most of the off-diagonal entries of $\mathbf{\tilde{B}}$ are zero, while the rest have negative values. These properties readily carry over to $\mathbf{B}$. 

A meaningful matrix factorization exploiting the aforementioned properties can be found by solving the problem
\begin{align}
\min_{\mathbf{B},\mathbf{S}} ~&~\tfrac{1}{2}\|\mathbf{B}\mathbf{L}-\mathbf{S}\|_F^2+\kappa_1\|\mathbf{O}\odot\mathbf{B}\|_1 +\kappa_2\|\mathbf{S}\|_1  \label{eq:P1}\tag{P1}\\
~&~ +\kappa_3\|\mathbf{S}\|_*-\kappa_4\log|\mathbf{B}|\nonumber\\ 
\textrm{s.to}~&~\mathbf{B}\succ \mathbf{0}, ~\mathbf{O}\odot\mathbf{B}\leq \mathbf{0}.\nonumber
\end{align}
where $\{\kappa_j\}_{j=1}^4$ are positive tuning parameters; and matrix $\mathbf{O}:=\mathbf{1}\mathbf{1}'-\mathbf{I}$ is introduced to select the off-diagonal entries of $\mathbf{B}$. The cost in \eqref{eq:P1} involves a least-squares data fitting term and four regularization components: $\|\mathbf{B}\|_1$ and $\|\mathbf{S}\|_1$ promote sparse $\mathbf{B}$ and $\mathbf{S}$ minimizers; $\|\mathbf{S}\|_*$ encourages a low-rank  solution for $\mathbf{S}$; while $-\log|\mathbf{B}|$ ensures $\mathbf{B}\succ \mathbf{0}$ and excludes the uninteresting case of having zero $(\mathbf{B},\mathbf{S})$ minimizers.

Problem \eqref{eq:P1} can be interpreted as a \emph{regularized maximum likelihood estimator} (MLE) of $(\mathbf{B},\mathbf{S})$. Suppose the noise term $\mathbf{n}_t$ in \eqref{eq:lmps2} is drawn from a multivariate Gaussian distribution with zero mean and covariance matrix $\sigma_n^2\mathbf{B}^{-2}$ for some $\sigma_n^2$. The negative log-likelihood of $\boldsymbol{\lambda}_t$ reads 
$-\log p(\boldsymbol{\lambda}_t;\mathbf{B},\sigma_n^2,\mathbf{s}_t)=\tfrac{1}{2\sigma_n^2}\|\mathbf{B}\boldsymbol{\lambda}_t-\mathbf{s}_t\|_2^2 -\log|\mathbf{B}|-\tfrac{N}{2}\log(2\pi\sigma_n^2)$.
Assuming independence across $\{\mathbf{n}_t\}_{t=1}^T$, the MLE of $(\mathbf{B},\mathbf{S})$ given the observations $\mathbf{L}$ is found as the solution of
\begin{equation}\label{eq:MLE}
\min_{\mathbf{B}\succ \mathbf{0},\mathbf{S}}~\tfrac{1}{2}\|\mathbf{B}\mathbf{L}-\mathbf{S}\|_F^2-T\sigma_n^2\log|\mathbf{B}|. 
\end{equation}
The extra penalties and the constraint in \eqref{eq:P1} regularize the MLE to further promote the structural properties of $(\mathbf{B},\mathbf{S})$.


\section{Algorithms }\label{sec:algorithms}
Optimization problem \eqref{eq:P1} is convex and it can be actually expressed as a semidefinite program (SDP). However, high-dimensional market data ($N$ and $T$ in the order of thousands), exclude the possibility of using standard interior point-based SDP solvers. Instead, an efficient algorithm based on the alternating direction method of multipliers (ADMM) is developed next; see e.g., \cite{Boyd10} for a review on ADMM. First, \eqref{eq:P1} is equivalently reformulated as
\begin{align}
\min_{{\mathbf{S}_1,\mathbf{S}_2 \atop \mathbf{B}_1,\mathbf{B}_2,\mathbf{B}_3}} &\tfrac{1}{2}\|\mathbf{B}_1\mathbf{L}{-}\mathbf{S}_2\|_F^2+\kappa_1\|\mathbf{O}\odot\mathbf{B}_2\|_1 + \kappa_2\|\mathbf{S}_1\|_1  \label{eq:P2}\tag{P2}\\
~&~+\kappa_3\|\mathbf{S}_2\|_*-\kappa_4\log|\mathbf{B}_3|\nonumber\\ 
\textrm{s.to}~&~\mathbf{O}\odot\mathbf{B}_2\leq \mathbf{0},~\mathbf{B}_3\succ \mathbf{0}\label{eq:P2a}\tag{P2a}\\
~&~\mathbf{B}_1=\mathbf{B}_2,~\mathbf{B}_1=\mathbf{B}_3, 
~\mathbf{S}_1=\mathbf{S}_2. \label{eq:P2b}\tag{P2b}
\end{align}
To efficiently handle the non-differentiable cost and the constraints in \eqref{eq:P1}, the original variables $\mathbf{B}$ and $\mathbf{S}$ are replaced by $(\mathbf{B}_1,\mathbf{B}_2,\mathbf{B}_3)$ and $(\mathbf{S}_1,\mathbf{S}_2)$ in \eqref{eq:P2}. Consensus among these variable duplicates is guaranteed by the constraints in \eqref{eq:P2b}.

If $\mathbf{Y}_{12}$, $\mathbf{Y}_{13}$, and $\mathbf{Y}$ are the Lagrange multipliers associated respectively with the three constraints in \eqref{eq:P2b}, the \emph{augmented} Lagrangian function of \eqref{eq:P2} is given by
\begin{align}\label{eq:augmented}
L_{\rho}&(\mathbf{B}_1,\mathbf{B}_2,\mathbf{B}_3,\mathbf{S}_1,\mathbf{S}_2;\mathbf{Y}_{12},\mathbf{Y}_{13},\mathbf{Y}):=\\
&\tfrac{1}{2}\|\mathbf{B}_1\mathbf{L}-\mathbf{S}_2\|_F^2+\kappa_1\|\mathbf{O}\odot\mathbf{B}_2\|_1 + \kappa_2\|\mathbf{S}_1\|_1 \nonumber\\
&+\kappa_3\|\mathbf{S}_2\|_* -\kappa_4\log|\mathbf{B}_3|+\mathbb{I}(\mathbf{O}\odot\mathbf{B}_2\leq\mathbf{0})  + \mathbb{I}(\mathbf{B}_3\succ \mathbf{0})  \nonumber\\
&+ \mathbf{Y}_{12}\bullet (\mathbf{B}_1-\mathbf{B}_2) + \mathbf{Y}_{13}\bullet (\mathbf{B}_1-\mathbf{B}_3) + \mathbf{Y}\bullet (\mathbf{S}_1-\mathbf{S}_2)\nonumber\\
&+ \tfrac{\rho}{2}\|\mathbf{B}_1-\mathbf{B}_2\|_F^2 + \tfrac{\rho}{2}\|\mathbf{B}_1-\mathbf{B}_3\|_F^2+\tfrac{\rho}{2}\|\mathbf{S}_1-\mathbf{S}_2\|_F^2\nonumber
\end{align}
where $\rho>0$ is a predefined constant and $\mathbb{I}(\mathcal{S})$ denotes the indicator function for set $\mathcal{S}$. Each ADMM iteration consists of two primal and one dual update steps. During a primal update step, $L_{\rho}$ is minimized over a subset of variables, while the remaining variables are fixed to their most recent values in a Gauss-Seidel fashion. In the dual update step, Lagrange multipliers are updated via gradient ascent. Letting $i$ denote the iteration index, our ADMM algorithm cycles through the three steps detailed next. 

In the \emph{first step}, $(\mathbf{B}_1,\mathbf{S}_1)$ are updated as the minimizers of $L_{\rho}(\mathbf{B}_1,\mathbf{B}_2^i,\mathbf{B}_3^i,\mathbf{S}_1,\mathbf{S}_2^i;\mathbf{Y}_{12}^i,\mathbf{Y}_{13}^i,\mathbf{Y}^i)$, where the superscript $i$ indicates the variable value at the end of the $i$-th iteration. The minimization is separable over $\mathbf{B}_1$ and $\mathbf{S}_1$. Regarding the update of $\mathbf{B}_1$, upon completing the squares and ignoring constant terms, $\mathbf{B}_1^{i+1}$ turns out to be the solution of
\begin{align}\label{eq:B1problem}
\min_{\mathbf{B}_1} ~\tfrac{1}{2}\|\mathbf{B}_1\mathbf{L}-\mathbf{S}_2^i\|_F^2+\rho\left\|\mathbf{B}_1 -\tfrac{\rho\mathbf{B}_2^i+\rho\mathbf{B}_3^i-\mathbf{Y}_{12}^i-\mathbf{Y}_{13}^i}{2\rho}\right\|_F^2
\end{align}
that is provided in closed form as
\begin{align*}
\mathbf{B}_1^{i+1}{:=}(\mathbf{S}_2^{i}\mathbf{L}' + \rho\mathbf{B}_2^i + \rho\mathbf{B}_3^i-\mathbf{Y}_{12}^i-\mathbf{Y}_{13}^i )(\mathbf{L}\mathbf{L}'+2\rho\mathbf{I})^{-1}.
\end{align*}
Optimization with respect to $\mathbf{S}_1$ entails solving
\begin{align}\label{eq:S1problem}
\min_{\mathbf{S}_1}~\tfrac{\rho}{2}\|\mathbf{S}_1-\mathbf{S}_2^i+\tfrac{1}{\rho}\mathbf{Y}^i\|_F^2+\kappa_2\|\mathbf{S}_1\|_1
\end{align}
whose minimizer becomes available in closed form as \cite{Boyd10}
\begin{align}\label{eq:S1}
\mathbf{S}_1^{i+1}:=\mathcal{S}_{\kappa_2/\rho}\left[\mathbf{S}_2^i - \tfrac{1}{\rho}\mathbf{Y}^i\right]
\end{align}
and $\mathcal{S}_{\alpha}\left[x\right]:=x\cdot \max\left\{1-\tfrac{\alpha}{|x|}, 0\right\}$ is the so termed soft thresholding operator applied entrywise to matrix $\mathbf{S}_2^i - \tfrac{1}{\rho}\mathbf{Y}^i$.

In the \emph{second step}, $(\mathbf{B}_2,\mathbf{B}_3,\mathbf{S}_2)$ are updated as the minimizers of $L_{\rho}(\mathbf{B}_1^{i+1},\mathbf{B}_2,\mathbf{B}_3,\mathbf{S}_1^{i+1},\mathbf{S}_2;\mathbf{Y}_{12}^i,\mathbf{Y}_{13}^i,\mathbf{Y}^i)$. The optimization is again separable over the three variables. Specifically, the update $\mathbf{B}_2^{i+1}$ is found as the solution of
\begin{align}\label{eq:B2problem}
\min_{\mathbf{O}\odot\mathbf{B}_2\leq \mathbf{0}}\tfrac{\rho}{2}\|\mathbf{B}_1^{i+1}-\mathbf{B}_2+\tfrac{1}{\rho}\mathbf{Y}_{12}^i\|_F^2+\kappa_1\|\mathbf{O}\odot\mathbf{B}_2\|_1.
\end{align}
Problem \eqref{eq:B2problem} decouples over the entries of $\mathbf{B}_2$ and after using the constraint to simplify the cost, it can be shown that
\begin{align*}
[\mathbf{B}_2^{i+1}]_{nm}{:=}\left\{\begin{array}{ll}
[\mathbf{B}_1^{i+1} + \tfrac{1}{\rho}\mathbf{Y}_{12}^i]_{nm},&n=m\\
\min\left\{[\mathbf{B}_1^{i+1} + \tfrac{1}{\rho}\mathbf{Y}_{12}^i]_{nm} + \tfrac{\kappa_1}{\rho},0\right\},&n\neq m
\end{array}\right..
\end{align*}
Next, variable $\mathbf{B}_3$ is updated as the minimizer of
\begin{align}\label{eq:B3problem}
\min_{\mathbf{B}_3\succ \mathbf{0}}~\tfrac{\rho}{2}\|\mathbf{B}_1^{i+1}-\mathbf{B}_3+\tfrac{1}{\rho}\mathbf{Y}_{13}^i\|_F^2 -\kappa_4\log|\mathbf{B}_3|.
\end{align}
By Lemma~\ref{le:logdet} (proved in the Appendix), the minimizer of \eqref{eq:B3problem} is neatly expressed as
\begin{align}\label{eq:B3}
\mathbf{B}_3^{i+1}=\mathcal{T}_{\kappa_4/\rho}\left[\mathbf{B}_1^{i+1}+\tfrac{1}{\rho}\mathbf{Y}_{13}^i\right]
\end{align}
where $\mathcal{T}_{\alpha}\left[\mathbf{X}\right]:= \mathbf{U}\diag\left(\left\{ \tfrac{1}{2}(s_k+\sqrt{s_k^2+4\alpha}) \right\}\right)\mathbf{U}'$ with the eigenvalue decomposition $\tfrac{1}{2}(\mathbf{X}+\mathbf{X}')=\mathbf{U}\diag(\{s_k\})\mathbf{U}'$. Finally, upon completing the squares, $\mathbf{S}_2^{i+1}$ is found via
\begin{align}\label{eq:S2problem}
\min_{\mathbf{S}_2} ~ \left\|\mathbf{S}_2-\tfrac{\mathbf{B}_1^{i+1}\mathbf{L}+ \rho\mathbf{S}_1^{i+1}+\mathbf{Y}^i}{\rho+1}\right\|_F^2+\tfrac{2\kappa_3}{\rho+1}\|\mathbf{S}_2\|_*
\end{align}
whose solution can be expressed as \cite[Th.~2.1]{SVT}
\begin{align}\label{eq:S2}
\mathbf{S}_2^{i+1}:=\tfrac{1}{\rho+1}\cdot\mathcal{P}_{\kappa_3}\left[\mathbf{B}_1^{i+1}\mathbf{L}+ \rho\mathbf{S}_1^{i+1}+\mathbf{Y}^i\right]
\end{align}
where $\mathcal{P}_{\alpha}\left[\mathbf{X}\right]:= \mathbf{U} \diag\left( \max\{\sigma_i-\alpha,0\}\right)\mathbf{V}'$, with the singular value decomposition $\mathbf{X}:=\mathbf{U}\diag\left(\{\sigma_i\}\right)\mathbf{V}'$. The closed-form update in \eqref{eq:S2} is essentially a soft thresholding operator on the singular values of its matrix argument. 

In the \emph{third step}, the Lagrange multipliers are updated via gradient ascent simply as
\begin{align}\label{eq:dual}
\mathbf{Y}_{12}^{i+1}&:= \mathbf{Y}_{12}^{i} + \rho(\mathbf{B}_1-\mathbf{B}_2)\\
\mathbf{Y}_{13}^{i+1}&:= \mathbf{Y}_{13}^{i} + \rho(\mathbf{B}_1-\mathbf{B}_3)\\
\mathbf{Y}^{i+1}&:= \mathbf{Y}^{i} + \rho(\mathbf{S}_1-\mathbf{S}_2).
\end{align}

\begin{table*}[t]
\renewcommand{\arraystretch}{1.00}
\vspace*{-1em}
\caption{Modeling parameters} \label{tbl:t} \centering
\vspace*{-1em}
\begin{tabular}{c|cccccccccccccc}
\hline
\hline
Bus number & 1 &2& 3& 4& 5&6&7&8&9&10&11&12&13&14\\ 
\hline
Type (Generator/Load/Zero)  & G(R) &G& G& L& L&G&Z&G&L&L&L&L&L&L\\ 
Unmodulated mean loads [MW]& - & 21.7 & 94.2 & 47.8 & 7.6 & 11.2 & - & - & 29.5 & 9.0 & 3.5 & 6.1 & 13.5& 14.9\\
 Mean generation bid [\$/MWh] & 18& 31 &30& - & -&15& - &22& - & - & -& -& -&-\\
 Upper generation bounds [MW] & 200 &140& 100& -& -&100&-&100&-&-&-&-&-&-\\ 
\hline
\hline
\end{tabular}
\vspace*{-1.5em}
\end{table*}

\section{Numerical Tests}\label{sec:simulations}
The novel topology recovery approach was evaluated using market data generated for the IEEE 14-bus grid. Ex-ante real-time prices were simulated over a one-day period by solving \eqref{eq:Bdcopf} for $288$ 5-min intervals using YALMIP and SDPT3 solvers~\cite{YALMIP}, \cite{SDPT3}. Matrices $\mathbf{B}$, $\mathbf{A}$, and $\mathbf{D}$ were obtained from MATPOWER~\cite{MATPOWER}. Among the 14 buses, buses $\{1,2,3,6,8\}$ are generators (bus 1 is the reference), bus 7 is a zero-injection bus, and the rest are inelastic loads. Lower generation bounds were set to zero and upper generation bounds were selected as listed in Table~\ref{tbl:t}. Following MATPOWER's line ordering, the flow limits $\overline{\mathbf{f}}$ on the $L=20$ transmission lines were set to $\{$70, 90, 50, 70, 50, 20, 50, 70, 90, 90, 20, 70, 50, 70, 20, 50, 90, 50, 50, 70$\}$MW.

Generation bids $\mathbf{c}$ were modeled as uniform random variables having the means shown in Table~\ref{tbl:t}, and standard deviation 2.88\$/MWh. Loads were drawn as independent Gaussian random variables with standard deviation $\sqrt{3}$MW, and means simulated as the product of the loads shown in Table~\ref{tbl:t} times a factor of $\{$0.77, 0.74, 0.73, 0.74, 0.77, 0.83, 0.90, 0.95, 0.97, 0.99, 0.99, 1.00, 0.99, 0.99, 0.97, 0.96, 0.94, 0.92, 0.91, 0.90, 0.91, 0.88, 0.81, 0.75$\}$ depending on the hour of the day. These factors were the normalized hourly loads over the Midwest Independent System Operator market in June 1st, 2012. After solving \eqref{eq:Bdcopf} for all 288 intervals, the price matrix was constructed as $\mathbf{B}^{-1}\mathbf{S}$ according to \eqref{eq:lmps2}-\eqref{eq:model}. Among the 288 intervals, congestion occured only in $T=168$ intervals; and these non-zero $\boldsymbol{\lambda}_t$'s comprised the columns of the nominal $13\times 168$ price matrix $\mathbf{L}_o$. 

Before inferring $(\mathbf{B},\mathbf{S})$ from \eqref{eq:P1}, the parameter vector $\boldsymbol{\kappa} :=[\kappa_1~\kappa_2~\kappa_3~\kappa_4]'$ needs to be tuned. Since cross-validation cannot be applied here~\cite{Hastie}, the following heuristic was used instead.
A grid of values was chosen for all $\kappa_j$'s. For each candidate $\boldsymbol{\kappa}$, 10\% of the entries of $\mathbf{L}_o$ chosen uniformly at random were considered unknown, and set to zero to yield the observed price matrix $\mathbf{L}$. 
Matrices $(\hat{\mathbf{B}},\hat{\mathbf{S}})$ were then found as the minimizers of \eqref{eq:P1} using $\mathbf{L}$. The entries of $\mathbf{L}_o$ assumed unknown were recovered as the corresponding entries of $\hat{\mathbf{B}}^{-1}\hat{\mathbf{S}}$. The process was repeated 10 times for each $\boldsymbol{\kappa}$. The squared reconstruction error on the missing entries of $\mathbf{L}_o$ was averaged over all 10 runs. The configuration attaining the lowest aggregate error was $\boldsymbol{\kappa}_o=[1{\cdot}10^{-3},~5{\cdot}10^{-4},~1{\cdot}10^{-2},~1{\cdot}10^{-1}]'$.

\begin{figure}
\vspace*{-0.5em}
\centering
\includegraphics[width=0.44\textwidth]{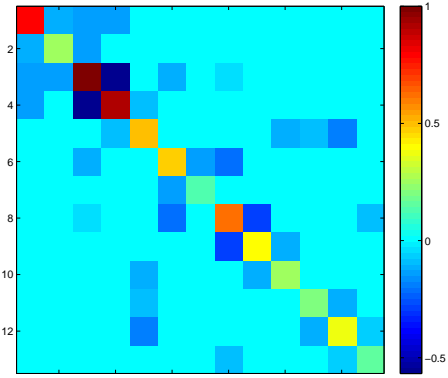}
\vspace*{-2.3em}
\caption{Nominal reduced Laplacian matrix $\mathbf{B}$ for the IEEE 14-bus grid $(N=13)$. Matrix $\mathbf{B}$ has been normalized to unit maximum value.}
\label{fig:Bnominal}
\vspace*{-1em}
\end{figure}

\begin{figure}
\vspace*{-0.2em}
\centering
\includegraphics[width=0.44\textwidth]{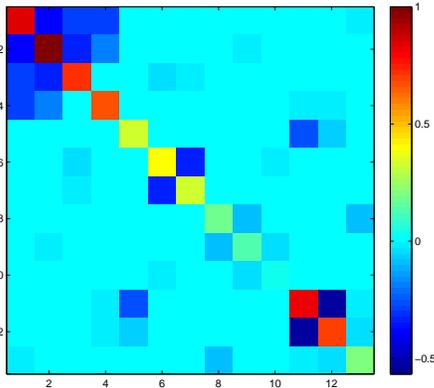}
\vspace*{-2.3em}
\caption{Reduced Laplacian matrix $\hat{\mathbf{B}}$ recovered by solving \eqref{eq:P1} using parameters $\boldsymbol{\kappa}$. Matrix $\hat{\mathbf{B}}$ has been normalized to unit maximum value.}
\label{fig:Brecovered}
\vspace*{-1em}
\end{figure}

Problem \eqref{eq:P1} was solved using the algorithm of Section~\ref{sec:algorithms}. Parameter $\rho$ was set to $10^3$, while the stopping criteria suggested in \cite[Sec.~3.3]{Boyd10} were employed. The ADMM solver terminated in less than 1 min on a 2.4 GHz Intel Core i7 processor with 4 GB RAM, whereas the SDPT3 solver could not run. Topology recovery results are shown in Figs.~\ref{fig:Bnominal}-\ref{fig:Brecovered}. The results are quite encouraging given that the novel scheme is based solely on the observed prices $\mathbf{L}$. Collecting prices over a longer observation period and over more diverse market conditions are expected to offer enhanced recovery.

\section{Conclusions}\label{sec:conclusions}
Recovering grid topology from LMPs was the theme of this work. It was first recognized that the price matrix admits an interesting bilinear decomposition. A convex optimization problem involving sparse and low-rank regularizers was formulated to reveal the constituent matrix factors. The problem was solved by an iterative ADMM-based algorithm entailing only closed-form updates. The novel scheme yielded encouraging topology recovery results on market data generated using the IEEE 14-bus grid. Performing identifiability analysis and experimenting with real market are challenging future directions.

\appendix\label{sec:appendix}
\begin{lemma}\label{le:logdet}
Given $\mathbf{A}\in\mathbb{R}^{N\times N}$ and $\alpha>0$, the convex problem $\min\left\{f(\mathbf{X}): \mathbf{X}\succeq \mathbf{0}\right\}$ where $f(\mathbf{X}):=\tfrac{1}{2} \|\mathbf{X}-\mathbf{A}\|_F^2 - \alpha \log |\mathbf{X}|$ admits the minimizer
\begin{equation}\label{eq:logdetmin}
\mathbf{X}_o:=\mathbf{U}\diag\left(\left\{ \tfrac{1}{2}(\lambda_k+\sqrt{\lambda_k^2+4\alpha)}{2} \right\}\right)\mathbf{U}'
\end{equation}
with the eigen-decomposition $\tfrac{1}{2}(\mathbf{A}+\mathbf{A}')=\mathbf{U}\diag(\{\lambda_k\})\mathbf{U}'$.
\end{lemma}

\begin{IEEEproof}
Consider the decomposition $\mathbf{A}=\mathbf{A}_s+\mathbf{A}_n$ into the symmetric $\mathbf{A}_s:=\tfrac{1}{2}(\mathbf{A}+\mathbf{A}')$ and the anti-symmetric $\mathbf{A}_n:=\tfrac{1}{2}(\mathbf{A}-\mathbf{A}')$. 
Recall that the inner product between a symmetric and an anti-symmetric matrix is zero.
For $\mathbf{X}_o$ to be the minimizer, it suffices to show that $(\nabla f(\mathbf{X}_o))\bullet (\mathbf{X}-\mathbf{X}_o)\geq 0$ for all $\mathbf{X}\succeq \mathbf{0}$. The gradient of $f(\mathbf{X})$ at $\mathbf{X}_o$ is $\nabla f(\mathbf{X}_o)=\mathbf{X}_o-\mathbf{A} - \alpha\mathbf{X}_o^{-1}=(\mathbf{X}_o-\mathbf{A}_s - \alpha\mathbf{X}_o^{-1})-\mathbf{A}_n$. Since $\mathbf{A}_n\bullet (\mathbf{X}-\mathbf{X}_o)=0$ for all $\mathbf{X}\succeq \mathbf{0}$, what remains to be shown is $(\mathbf{X}_o-\mathbf{A}_s - \alpha\mathbf{X}_o^{-1})\bullet (\mathbf{X}-\mathbf{X}_o)\geq 0$ for all $\mathbf{X}\succeq \mathbf{0}$. Observing that $\mathbf{X}_o$, $\mathbf{A}_s$, and $\alpha\mathbf{X}_o^{-1}$ share the same eigenvectors $\mathbf{U}$, it is easy to show that $\mathbf{X}_o-\mathbf{A}_s - \alpha\mathbf{X}_o^{-1}=\mathbf{0}$, which completes the proof.
\end{IEEEproof}

\bibliographystyle{IEEEtran}
\bibliography{IEEEabrv,power}
\end{document}